\documentclass[10pt,twocolumn,letterpaper]{article}

\usepackage[pagenumbers]{cvpr}

\usepackage{cuted} 

\usepackage{currfile} 

\usepackage{caption} 

\usepackage{algorithm}
\usepackage{algpseudocode}
\usepackage{float}
\usepackage{colortbl}
\usepackage{placeins}
\usepackage{flafter}

\definecolor{cvprblue}{rgb}{0.21,0.49,0.74}
\definecolor{gray20}{gray}{0.90}
\definecolor{gray10}{gray}{0.95}
\definecolor{highlightcyan}{HTML}{D7F6FF}
\usepackage[pagebackref,breaklinks,colorlinks,allcolors=cvprblue]{hyperref}

\def\confName{CVPR}
\def\confYear{2027}

\title{One Patch, Many Policies: Object-Associated Physical Patch Attacks on Robotic Manipulation}
\author{%
  Yiqian Liu\quad Yuqi Lin\quad Hui Lu\quad Jiaming He\quad Qi Zhu\\
  Deyi Ji\quad Bing Li\quad Kaizheng Wang\quad Zhiyuan Zha\quad Yi Yu\thanks{Corresponding author.}%
}
\date{}

\begin{document}
\maketitle

\begin{abstract}
Vision-Language-Action (VLA) and World Action Models (WAMs) are vulnerable to adversarial visual interference, motivating the study of universal and transferable physical patch attacks. We propose \textbf{Object-Associated Patch Attack (OAPA)}, a universal and transferable physical patch framework that \textbf{enables an attacker to select which task object’s manipulation to disrupt through patch placement}, without re-optimizing the patch. Our framework combines object-associated placement, differentiable physical compositing, and final action-expert representation divergence. Physical compositing accounts for perspective and occlusion across camera views, while representation divergence guides patch optimization using either relative RMS or cosine distance. Experimental results demonstrate strong attack transferability across task suites and policies, together with the ability to selectively disrupt manipulation of an attacker-chosen task object through patch placement.
\end{abstract}

\begin{figure*}[!t]
  \centering
  \includegraphics[width=\textwidth]{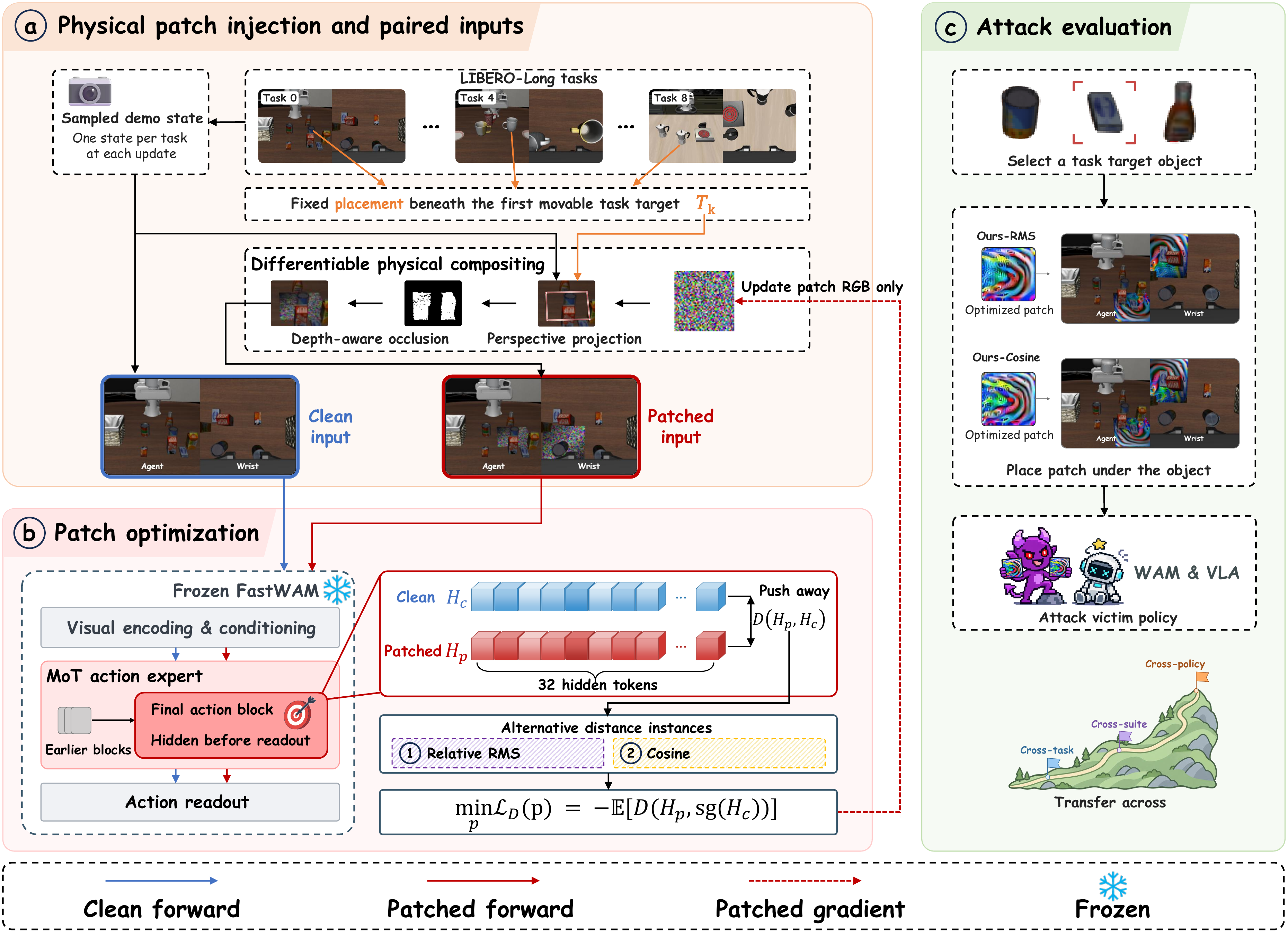}
  \vspace{-4mm}
  \caption{\textbf{Overview of our object-associated physical patch attack framework.} \textbf{(a)} During optimization, a shared patch is placed beneath the first movable task target at a fixed task-specific pose. Perspective projection and depth-aware occlusion produce paired clean and patched inputs from the agent and wrist views. \textbf{(b)} With FastWAM frozen, we optimize patch RGB values to maximize the discrepancy between all 32 final action-expert hidden tokens before action readout. Relative RMS and cosine distances define two independently optimized instances, Ours-RMS and Ours-Cosine. \textbf{(c)} Each optimized patch is deployed at a location associated with a selected task target without re-optimization and evaluated through closed-loop rollouts across tasks, suites, and victim policies.}
  \label{fig:method_overview}
  \vspace{-4mm}
\end{figure*}

\section{Introduction}
\label{sec:introduction}

Vision-Language-Action (VLA) and World Action Models (WAMs) enable robots to interpret visual observations and language instructions, generate actions, and perform diverse manipulation tasks~\cite{lit_2406_09246,black2024pi0,lit_2307_15818,yuan2026fastwam,chen2026lawam,kim2026cosmos}. Their growing capabilities support generalization across tasks and environments, while their reliance on visual inputs exposes a practical attack surface: adversarial changes to scene appearance can corrupt the representations used for action generation and disrupt closed-loop execution~\cite{lu2026patch,lit_2610_06814,lit_2411_13587}. These disturbances can propagate into physical actions, potentially causing failed manipulation, unintended motion, or unsafe interactions. Motivated by these risks, we conduct a systematic study of universal and transferable physical adversarial patches for VLA- and WAM-driven robots, with the objective of disrupting the manipulation of attacker-selected task objects.

Despite recent progress, achieving effective and controllable attacks under practical physical constraints remains challenging. Patches optimized against a particular policy or task distribution may exploit source-specific sensitivities, limiting their effectiveness on unseen policies and tasks~\cite{lu2026patch,lit_2610_06814,lit_2610_39816,lit_2510_13237}. Image-space patch formulations provide flexible visual perturbations, but their deployment assumptions do not necessarily preserve perspective and visibility consistency across camera views. Geometry-aware texture attacks improve physical consistency by attaching adversarial appearance to object surfaces, yet this also couples deployment to the selected carrier geometry and surface parameterization~\cite{lit_2608_13453,lit_2604_01618}. Moreover, strong task-level disruption does not by itself establish control over which task object's manipulation is disrupted. Specifying an erroneous action direction or sequence also differs from selecting an object-level failure outcome. The challenge is therefore to construct a reusable physical patch that combines strong attack effectiveness and black-box transfer with the ability to designate a task object whose manipulation is disrupted, without requiring a surface texture tailored to that object.

To address these challenges, we associate patch placement with task objects and optimize a shared patch through action-expert representations. Our placement analysis shows that object-associated locations yield stronger disruption than object-independent tabletop locations. The attacker can select another target by relocating the same patch without re-optimization. To connect optimization to action generation without prescribing task-specific erroneous actions, we target the final action-expert representations before action readout in a frozen surrogate policy. This allows joint patch optimization across tasks requiring different manipulation actions, promoting attack generalization. Differentiable physical compositing models perspective projection and depth-aware occlusion to maintain consistency across camera views. Together, these components form Object-Associated Patch Attack (OAPA), a universal and transferable physical patch framework integrating spatial target selection, action-expert representation divergence, and physically consistent compositing.

Our contributions are summarized as follows:
\begin{itemize}
    \item We introduce an object-targeted attack that uses patch placement to designate the task object whose manipulation is disrupted. The same universal patch can switch attack targets by being placed beneath different task objects, without changing its content or requiring further optimization.
    \item We propose OAPA, a physical patch attack framework that integrates spatial target selection, physically consistent dual-view compositing, and cross-task patch optimization. By modeling perspective projection and depth-aware occlusion, the framework preserves the patch's spatial and visibility consistency across both camera views.
    \item To the best of our knowledge, we are the first to optimize universal adversarial patches against the final hidden tokens of a frozen source policy's action expert, immediately before action readout. We instantiate representation divergence using relative RMS or cosine distance, enabling joint optimization across tasks requiring different manipulation actions without prescribing erroneous action sequences.
    \item We conduct extensive closed-loop evaluations across four task suites and five robot policies, demonstrating strong attack effectiveness and black-box transferability. Target-object switching and placement ablations further establish the role of object association and show that the disrupted object can be selected through patch placement.
\end{itemize}

\section{Related Work}
\label{sec:related_work}

\noindent\textbf{Vision-Language-Action and World Action Models.} Vision--language representation learning provides a foundation for connecting visual perception and language understanding with robotic control~\cite{lit_2303_03378,lit_2103_00020,lit_2303_15343,lit_2402_07865}. Existing VLA policies generate actions through autoregressive prediction of discrete tokens or continuous sequence generation based on diffusion and flow matching~\cite{lit_2406_09246,black2024pi0,lit_2307_15818,lit_2504_16054,lit_2303_04137,lit_2410_07864}. Large-scale robot demonstrations and generalist policy learning further support generalization across tasks and embodiments~\cite{lit_2212_06817,lit_2405_12213,lit_2310_08864,lit_2403_12945,lit_2308_12952}. Beyond directly mapping observations to actions, predictive world modeling has also been incorporated into robot policy learning. World action models exploit visual or latent predictions of scene evolution through different forms of coupling between world modeling and action generation~\cite{yuan2026fastwam,chen2026lawam,kim2026cosmos,lit_2608_04996,lit_2512_13030,lit_2602_15922,lit_2608_05903}. These complementary developments motivate examining whether adversarial vulnerabilities transfer across policies with different representations and action-generation mechanisms.

\noindent\textbf{Adversarial Attacks on Robot Policies.} Adversarial research considers both white-box access and black-box settings, with cross-model transfer enabling attacks without victim-model gradients~\cite{lit_1312_6199,lit_1412_6572,lit_1706_06083,lit_1608_04644,lit_1511_04599,lit_1610_08401,lit_1611_02770,wang2025breaking}. Transfer-oriented approaches improve generalization through gradient optimization, input transformations, and intermediate-representation objectives~\cite{lit_1710_06081,lit_1803_06978,lit_1904_02884,lit_1908_06281,lit_2103_15571,lit_2102_00436,lit_1907_10823,lit_2204_00008,lit_2002_05990,zhou2023advclip,zhou2023downstream}, while universal perturbations and physical patches extend attacks across inputs and viewing conditions~\cite{lit_1712_09665,lit_1707_07397,lit_1707_08945,lit_1904_08653,lit_1804_05810,lit_1610_08401}. In robotics, language manipulation, visual perturbations, and model backdoors expose different vulnerabilities in instruction-conditioned control~\cite{lit_2411_13587,lit_2511_16203,lit_2410_13691,lit_2407_20242,lit_2506_03350,lit_2505_16640,wang2025advedm}. Within visual attacks, existing approaches exploit action-space objectives, representation discrepancies, and attention-related signals to disrupt execution and promote transfer~\cite{lu2026patch,lit_2610_06814,lit_2608_03207,lit_2610_39816,lit_2411_13587,lit_2606_03556,lit_2610_09708,lit_2605_28083,lit_2510_13237}. Geometry-aware appearance attacks further account for object surfaces and rendering during adversarial optimization~\cite{lit_2608_13453,lit_2604_01618,yan2026vfacamou}. Existing targeted attacks can steer policy outputs toward attacker-specified actions~\cite{lit_2608_13453}. Such action-level targeting differs from selecting the task object whose manipulation is disrupted: one specifies how the policy should deviate, while the other specifies which object should be affected. Overall task failure alone does not establish object-level selectivity. Accordingly, our work investigates object-level targeted attacks on VLA- and WAM-driven robots, where the same universal, transferable physical patch selects the task object to disrupt through its placement.

\section{Preliminaries and Problem Formulation}
\subsection{Preliminaries}
\noindent\textbf{Robot policies.} We consider a robot performing a language-conditioned manipulation task $k$ from a scene state $s$. Let $I_v(s)$ denote the RGB observation captured by camera $v\in\{\mathrm{agent},\mathrm{wrist}\}$. The clean dual-view observations are
\begin{equation}
\mathbf I_c(s)=\bigl(I_{\mathrm{agent}}(s),I_{\mathrm{wrist}}(s)\bigr).
\label{eq:clean_dual_observations}
\end{equation}
Given the robot proprioception $q_s$, task instruction $c_k$, and initial action noise $\eta$, our chosen surrogate policy, FastWAM~\cite{yuan2026fastwam}, predicts an action chunk
\begin{equation}
A=f_\theta\bigl(\mathbf I_c(s),q_s,c_k;\eta\bigr)\in\mathbb R^{32\times7},
\end{equation}
where $\theta$ denotes the policy parameters. Here, $A$ contains 32 seven-dimensional control actions.

\noindent\textbf{Physical adversarial patch.} We represent a shared physical patch by its RGB values $p\in\mathcal P=[0,1]^{50\times50\times3}$, mapped onto a square plane with a physical side length of $0.25\,\mathrm m$. Let $T\in\mathbb R^{4\times4}$ denote a rigid transformation comprising rotation and translation that maps patch-local coordinates to world coordinates, and let $\xi$ denote a sampled patch transformation. For a scene state $s$, the patched dual-view observations are
\begin{equation}
\mathbf I_p=\mathcal C(s,p,T,\xi),
\end{equation}
where $\mathcal C$ projects the same transformed physical patch into both camera views and determines its visibility using scene depth. Here, $p$ specifies the patch appearance, $T$ its placement, and $\xi$ its sampled transformation. The projection and compositing operations are defined in Sec.~\ref{sec:physical_rendering}.

\subsection{Problem Formulation}
\label{sec:formulation}
\label{sec:overview}
\noindent\textbf{Object-associated placement.} Let $\mathcal O_k$ denote the set of task-specified target objects in task $k$, and let $s_k^0$ be its official reference initial state (state 0). For an object $o\in\mathcal O_k$, we define its associated patch pose as
\begin{equation}
T_{k,o}=\operatorname{PlaceAssociated}(o,s_k^0),
\label{eq:task_patch_pose}
\end{equation}
where $\operatorname{PlaceAssociated}$ assigns a task-specific patch pose at a location spatially associated with the selected target object. During optimization, we use the ten tasks in LIBERO-Long, denoted by $\mathcal K_{\mathrm{opt}}$, and fix the placement beneath the first movable task-specified target object:
\begin{equation}
\begin{aligned}
o_k^{\mathrm{opt}}&=\operatorname{FirstMovableTaskObject}(k),\\
T_k&=T_{k,o_k^{\mathrm{opt}}}=\operatorname{PlaceBelow}(o_k^{\mathrm{opt}},s_k^0).
\end{aligned}
\label{eq:optimization_patch_pose}
\end{equation}
Here, $\operatorname{FirstMovableTaskObject}(k)$ selects the first movable target object in the order in which target objects are mentioned in the task instruction $c_k$, and $\operatorname{PlaceBelow}$ places the patch on the supporting surface beneath the object's reference position.
For each task, $T_k$ remains unchanged across sampled demonstration states. At deployment, the attacker may select any $o\in\mathcal O_k$ and place the optimized patch at $T_{k,o}$, without modifying its content or performing further optimization. The selected patch pose remains fixed in world coordinates throughout the rollout.

\noindent\textbf{Threat model.} The attacker has white-box access to a frozen surrogate policy and access to the simulator observations, camera parameters, scene depth, and reference object poses required for physical compositing. Only the patch RGB values $p$ are optimized; the policy parameters, task instructions, and object dynamics remain unchanged. The optimized patch is shared across tasks and camera views and transferred to victim policies without using their parameters, gradients, or rollout feedback for optimization. The attack aims to disrupt manipulation of an attacker-selected task object and thereby reduce closed-loop task success under the original instruction.

\noindent\textbf{Optimization objective.} Let $\mathcal D_k$ denote the demonstration-state sampling distribution for task $k$. At each optimization step, we sample one demonstration state $s_k\sim\mathcal D_k$ for each task $k\in\mathcal K_{\mathrm{opt}}$, together with a patch transformation $\xi_k$ and initial action noise $\eta_k$, and average the losses equally across tasks. For a fixed feature-distance choice $D\in\{D_R,D_C\}$, we optimize a shared patch by
\begin{equation}
\begin{aligned}
p_D^\star&\in\arg\min_{p\in\mathcal P}\mathcal L_D(p),\\
\mathcal L_D(p)&=\mathbb E_{k,s,\xi,\eta}\!\left[\ell_D(p;k,s,\xi,\eta)\right].
\end{aligned}
\label{eq:patch_objective}
\end{equation}
The expectation assigns equal weight to tasks in $\mathcal K_{\mathrm{opt}}$. Here, $D_R$ and $D_C$ denote the two distance instances implemented in our framework: relative RMS distance and cosine distance, respectively. The loss $\ell_D$ measures patch-induced representation discrepancy in the frozen surrogate policy and is constructed in Sec.~\ref{sec:action_token_divergence}. The task-specific placement $T_k$ enters the loss through the patched observations $\mathcal C(s,p,T_k,\xi)$. Each distance defines an independent optimization instance, producing the Ours-RMS or Ours-Cosine patch. Attack effectiveness is assessed by the closed-loop task success rate under the original task instruction.

\section{Methodology}
\label{sec:method}

\subsection{Object-Associated Physical Compositing}
\label{sec:physical_rendering}

We implement the compositing map $\mathcal C(s,p,T_k,\xi)$ by transforming a shared physical patch, projecting it into both camera views, and resolving visibility against scene depth. The task-specific pose $T_k$ is defined in Sec.~\ref{sec:formulation}; the following operations are shared by both distance instances.

\noindent\textbf{Shared patch transformation.} We parameterize the sampled transformation as $\xi=(\alpha,\beta,\gamma)$, where the components control in-plane rotation, brightness, and contrast, respectively. Let $\widetilde R(\alpha)\in\mathbb R^{4\times4}$ denote the homogeneous rotation about the patch-local normal through its center. The transformed pose is
\begin{equation}
T_{k,\xi}^{\mathrm{opt}}=T_k\widetilde R(\alpha).
\label{eq:transformed_patch_pose}
\end{equation}
This rotation preserves the patch center, supporting plane, and physical size. The appearance transformation $\mathcal A_{\beta,\gamma}(p)$ adjusts brightness and contrast without changing the geometry. The same sampled transformation is used for both views, implementing expectation over transformation (EOT) on a shared scene patch.

\noindent\textbf{Dual-view perspective projection.} Let $u_i\in\mathbb R^2$, $i=1,\ldots,4$, denote the patch corners in local metric coordinates, centered at the origin and spanning the physical side length of $0.25\,\mathrm m$. With homogeneous coordinates $\bar u_i=(u_i^\top,0,1)^\top$, the world coordinates and image projections are
\begin{equation}
\begin{aligned}
\bar X_i&=T_{k,\xi}^{\mathrm{opt}}\bar u_i,\\
x_{v,i}&=\pi\!\left(P_v(s)\bar X_i\right),\quad v\in\{\mathrm{agent},\mathrm{wrist}\},
\end{aligned}
\label{eq:patch_projection}
\end{equation}
where $P_v(s)$ is the camera projection matrix at state $s$, and $\pi$ converts homogeneous image coordinates to two-dimensional coordinates by perspective division. The projected corners define a quadrilateral $Q_v$ and a homography $G_v$ mapping patch texture coordinates to image coordinates. For $x\in Q_v$, the projected patch color is
\begin{equation}
W_v(x)=\operatorname{Bilinear}\!\left(\mathcal A_{\beta,\gamma}(p),\pi(G_v^{-1}\bar x)\right),
\label{eq:warped_patch}
\end{equation}
where $\bar x=(x^\top,1)^\top$, and $\operatorname{Bilinear}$ samples the transformed patch at the inverse-mapped texture coordinates. Thus, the two views observe camera-dependent projections of the same physical patch.

\noindent\textbf{Depth-aware compositing.} For each $x\in Q_v$, let $z_{p,v}(x)$ and $z_{s,v}(x)$ denote the patch-plane and clean-scene depths, expressed in the same camera depth convention. The visibility mask is
\begin{equation}
M_v(x)=\mathbf 1\!\left[z_{p,v}(x)\le z_{s,v}(x)+\delta\right],
\label{eq:patch_visibility}
\end{equation}
where $\mathbf 1[\cdot]$ is the indicator function and $\delta$ is a depth tolerance. We form the patched image as
\begin{equation}
I_{p,v}(x)=
\begin{cases}
(1-M_v(x))I_v(s)(x)+M_v(x)W_v(x),&x\in Q_v,\\
I_v(s)(x),&x\notin Q_v.
\end{cases}
\label{eq:physical_compositor}
\end{equation}
Foreground objects therefore retain their original pixels wherever they occlude the patch. The resulting dual-view observations are
\begin{equation}
\mathbf I_p=\bigl(I_{p,\mathrm{agent}},I_{p,\mathrm{wrist}}\bigr)=\mathcal C(s,p,T_k,\xi).
\label{eq:paired_observations}
\end{equation}
For fixed $s,T_k,\xi$, the projection and visibility masks are independent of the RGB parameters $p$. Gradients from visible patch pixels propagate through bilinear sampling to the shared patch, allowing optimization through both camera views.

\begin{figure}[!t]
\begingroup
\hrule height 0.5pt
\nobreak\vspace{2pt}
\captionsetup{hypcap=false,labelformat=empty,labelfont=bf,textfont=md,labelsep=space,justification=raggedright,singlelinecheck=false}
\captionof{algorithm}[Optimization of the Object-Associated Patch]{\textbf{Algorithm~\thealgorithm}\enspace Optimization of the Object-Associated Patch}
\label{alg:patch_optimization}
\nobreak
\hrule height 0.4pt
\nobreak\vspace{2pt}
\algrenewcommand\alglinenumber[1]{\normalfont\scriptsize #1:}
\begin{algorithmic}[1]
\State \settowidth{\dimen0}{\textbf{Input:}\ }\hangindent=\dimen0\hangafter=1\relax
\textbf{Input:} Frozen token extractor $F_\theta$, tasks $\mathcal K_{\mathrm{opt}}$, demonstration-state distributions $\mathcal D_k$, reference states $s_k^0$, update budget $U$, and distance $D\in\{D_R,D_C\}$
\State Set $B=|\mathcal K_{\mathrm{opt}}|$ and index the tasks as $(k_b)_{b=1}^B$
\State Fix each $T_k$ beneath $o_k^{\mathrm{opt}}$ using Eq.~\eqref{eq:optimization_patch_pose}
\State Initialize $p_0\in\mathcal P$ and initialize Adam
\For{$u=1,\ldots,U$}
  \For{$b=1,\ldots,B$}
    \State Sample $s_b\sim\mathcal D_{k_b}$, $\xi_b$, and $\eta_b$
    \State Compute $T_{k_b,\xi_b}^{\mathrm{opt}}$ using Eq.~\eqref{eq:transformed_patch_pose}
    \State Form $\mathbf I_{c,b}=\mathbf I_c(s_b)$ via Eq.~\eqref{eq:clean_dual_observations}
\begin{samepage}
    \State Form $\mathbf I_{p,b}=\mathcal C(s_b,p_{u-1},T_{k_b},\xi_b)$
    \Statex \hspace{\algorithmicindent}\hspace{\algorithmicindent} using Eqs.~\eqref{eq:patch_projection}--\eqref{eq:paired_observations}
\end{samepage}
\begin{samepage}
    \State Use $a_{k_b,s_b,\eta_b}$ for both passes and extract
    \Statex \hspace{\algorithmicindent}\hspace{\algorithmicindent} $H_{c,b},H_{p,b}$ via Eq.~\eqref{eq:paired_action_tokens}
\end{samepage}
  \EndFor
  \State Evaluate $\widehat{\mathcal L}_{D,B}(p_{u-1})$ via Eq.~\eqref{eq:batch_patch_loss}
\begin{samepage}
  \State Apply Adam to $p_{u-1}$ using $\nabla_p\widehat{\mathcal L}_{D,B}(p_{u-1})$
  \Statex \hspace{\algorithmicindent}\hspace{\algorithmicindent} and clip to $\mathcal P$ to obtain $p_u$
\end{samepage}
\EndFor
\State \Return $p_D^\star=p_U$
\end{algorithmic}
\nobreak\hrule height 0.5pt
\endgroup
\end{figure}

\subsection{Final Action-Token Divergence}
\label{sec:action_token_divergence}

\noindent\textbf{Action-generation interface.} We target the final action-expert hidden tokens immediately before action readout in the frozen surrogate policy, FastWAM. Let $F_\theta$ extract these tokens, and abbreviate the shared conditioning as $a_{k,s,\eta}=(q_s,c_k,\eta)$. For the clean observations defined in Section 3.1 and patched observations constructed in Sec.~\ref{sec:physical_rendering}, we obtain
\begin{equation}
\begin{aligned}
H_c&=F_\theta\bigl(\mathbf I_c(s),a_{k,s,\eta}\bigr),\\
H_p&=F_\theta\bigl(\mathcal C(s,p,T_k,\xi),a_{k,s,\eta}\bigr),
\end{aligned}
\label{eq:paired_action_tokens}
\end{equation}
where $H_c,H_p\in\mathbb R^{32\times d_h}$, and $d_h$ is the hidden dimension. Here, $F_\theta(\mathbf I,a_{k,s,\eta})$ abbreviates $F_\theta(\mathbf I,q_s,c_k;\eta)$; both branches use the same proprioception, instruction, and initial action noise.

Let $\mathcal R_\theta(\cdot\,;a_{k,s,\eta})$ denote the subsequent action readout and denoising update:
\begin{equation}
\begin{aligned}
A_c&=\mathcal R_\theta(H_c;a_{k,s,\eta}),\\
A_p&=\mathcal R_\theta(H_p;a_{k,s,\eta}).
\end{aligned}
\label{eq:paired_action_readout}
\end{equation}
With the shared conditioning $a_{k,s,\eta}$ fixed, the complete policy is expressed as
\begin{equation}
f_\theta(\mathbf I,q_s,c_k;\eta)=\mathcal R_\theta\bigl(F_\theta(\mathbf I,a_{k,s,\eta});a_{k,s,\eta}\bigr).
\label{eq:policy_feature_readout}
\end{equation}
Define $h=\operatorname{vec}(H_c)$, $h_p=\operatorname{vec}(H_p)$, and $\Delta h=h_p-h$, where $\operatorname{vec}$ flattens a tensor into a vector. Under fixed conditioning, the local action response is
\begin{equation}
\operatorname{vec}(A_p-A_c)
=J_{\mathcal R}(h;a_{k,s,\eta})\Delta h+o(\|\Delta h\|_2),
\label{eq:local_action_response}
\end{equation}
where $J_{\mathcal R}$ is the Jacobian of the vectorized action output with respect to the vectorized hidden tokens, and the final term is the higher-order remainder. This relation connects perturbations at the selected representation to action generation. We therefore optimize representation discrepancy at this interface without prescribing a task-specific erroneous action sequence.

\noindent\textbf{Feature-distance instances.} We compute both distances over the complete tensor of 32 hidden tokens. For a tensor $X$ containing $n$ elements, define the stabilized RMS as
\begin{equation}
r_\epsilon(X)=\sqrt{\frac{\|X\|_F^2}{n}+\epsilon},
\end{equation}
where $\|\cdot\|_F$ is the Frobenius norm. The two distance instances are
\begin{equation}
\begin{aligned}
D_R(H_p,H_c)&=\frac{r_\epsilon(H_p-H_c)}{r_\epsilon(H_c)+\epsilon},\\
D_C(H_p,H_c)&=1-\cos_\epsilon(H_p,H_c),
\end{aligned}
\label{eq:hidden_divergence}
\end{equation}
where $\cos_\epsilon$ denotes cosine similarity between the flattened tensors with norm stabilization, and $\epsilon=10^{-8}$.

For nonzero $h,h_p$, let $\rho=\|h_p\|_2/\|h\|_2$, and let $\varphi$ be their angle. Omitting numerical stabilization gives
\begin{equation}
\begin{aligned}
D_R^2&=\frac{\|h_p-h\|_2^2}{\|h\|_2^2}=(\rho-1)^2+2\rho(1-\cos\varphi),\\
D_C&=1-\cos\varphi.
\end{aligned}
\label{eq:relative_rms_geometry}
\end{equation}
Thus, relative RMS measures overall displacement relative to the clean representation's scale, incorporating both norm and angular changes. Cosine distance measures angular deviation and is invariant to positive rescaling in this unstabilized form.

\begin{table*}[!t]
\centering
\caption{\textbf{Closed-loop task success across suites and policies.}
Each suite contributes 100 episodes per policy and condition. Values are percentages. Average gives equal weight to the four suites. Higher clean SR indicates stronger unperturbed performance, and lower SR under a patch indicates stronger disruption. Ours-RMS and Ours-Cosine use relative RMS and cosine distance, respectively.}
\vspace{-3mm}
\label{tab:canonical_transfer}
\small
\setlength{\tabcolsep}{4pt}
\renewcommand{\arraystretch}{0.80}
\resizebox{0.92\textwidth}{!}{%
  \begin{tabular}{llrrrrrrrr}
\toprule
\rowcolor{gray20}
Policy & Suite
& Clean $\uparrow$
& Random $\downarrow$
& UPA-RFAS $\downarrow$
& TAPDreamer $\downarrow$
& DRIFT $\downarrow$
& PASA-KV $\downarrow$
& \begin{tabular}[c]{@{}c@{}}Ours\\RMS\end{tabular}\,$\downarrow$
& \begin{tabular}[c]{@{}c@{}}Ours\\Cosine\end{tabular}\,$\downarrow$ \\
\midrule
 & Long & 93 & 42 & 19 & 24  & 11&19& \cellcolor{highlightcyan}9 & \cellcolor{highlightcyan}11 \\
 & Spatial & 98 & 55 & 54 & 43  & 57&39& \cellcolor{highlightcyan}58 & \cellcolor{highlightcyan}49 \\
FastWAM & Object & 96 & 69 & 42 & 27  & 29&21& \cellcolor{highlightcyan}10 & \cellcolor{highlightcyan}8 \\
 & Goal & 95 & 83 & 80 & 72  & 66&65& \cellcolor{highlightcyan}64 & \cellcolor{highlightcyan}62 \\
 & \textbf{Average} & \textbf{95.50} & \textbf{62.25} & \textbf{48.75} & \textbf{41.50}  & \textbf{40.75}&\textbf{36.00}& \cellcolor{highlightcyan}\textbf{35.25} & \cellcolor{highlightcyan}\textbf{32.50} \\
\midrule
 & Long & 98 & 82 & 71 & 77  & 36&54& \cellcolor{highlightcyan}22& \cellcolor{highlightcyan}19\\
 & Spatial & 100 & 97 & 93 & 97  & 70&93& \cellcolor{highlightcyan}65& \cellcolor{highlightcyan}63\\
LaWAM & Object & 100 & 100 & 79 & 93  & 42&76& \cellcolor{highlightcyan}20& \cellcolor{highlightcyan}16\\
 & Goal & 98 & 90 & 81 & 87  & 76&81& \cellcolor{highlightcyan}60& \cellcolor{highlightcyan}49\\
 & \textbf{Average} & \textbf{99.00} & \textbf{92.25} & \textbf{81.00} &\textbf{88.50}& \textbf{56.00}&\textbf{76.00}& \cellcolor{highlightcyan}\textbf{41.75}& \cellcolor{highlightcyan}\textbf{36.75}\\
\midrule
 & Long & 94 & 68 & 52 & 60  & 14&44& \cellcolor{highlightcyan}10 & \cellcolor{highlightcyan}10 \\
 & Spatial & 100 & 89 & 91 & 89  & 83&87& \cellcolor{highlightcyan}82 & \cellcolor{highlightcyan}74 \\
Cosmos Policy & Object & 100 & 100 & 78 & 97  & 21&95& \cellcolor{highlightcyan}8 & \cellcolor{highlightcyan}7 \\
 & Goal & 100 & 87 & 77 & 82  & 62&79& \cellcolor{highlightcyan}65 & \cellcolor{highlightcyan}64 \\
 & \textbf{Average} & \textbf{98.50} & \textbf{86.00} & \textbf{74.50} & \textbf{82.00}  & \textbf{45.00}&\textbf{76.25}& \cellcolor{highlightcyan}\textbf{41.25} & \cellcolor{highlightcyan}\textbf{38.75} \\
\midrule
 & Long & 92 & 71 & 40 & 67  & 8  &  58  & \cellcolor{highlightcyan}12 & \cellcolor{highlightcyan}10 \\
 & Spatial & 97 & 88 & 85 & 89  & 81  &  88  & \cellcolor{highlightcyan}79 & \cellcolor{highlightcyan}80 \\
OpenVLA-OFT & Object & 99 & 97 & 60 & 93  & 50&  87& \cellcolor{highlightcyan}35 & \cellcolor{highlightcyan}36 \\
 & Goal & 97 & 94 & 80 & 90  & 82  &92& \cellcolor{highlightcyan}71 & \cellcolor{highlightcyan}74 \\
 & \textbf{Average} & \textbf{96.25} & \textbf{87.50} & \textbf{66.25} & \textbf{84.75}  & \textbf{55.25}&\textbf{81.25}& \cellcolor{highlightcyan}\textbf{49.25} & \cellcolor{highlightcyan}\textbf{50.00} \\
\midrule
 & Long & 85 & 44 & 19 & 46  & 13  &  52  & \cellcolor{highlightcyan}3 & \cellcolor{highlightcyan}7 \\
 & Spatial & 93 & 91 & 62 & 80  & 49  &  88  & \cellcolor{highlightcyan}37 & \cellcolor{highlightcyan}44 \\
$\pi_0$ & Object & 99 & 86 & 66 & 87 & 36  &  82  & \cellcolor{highlightcyan}21 & \cellcolor{highlightcyan}20 \\
 & Goal & 95 & 84 & 49 & 85  & 41  &  69  & \cellcolor{highlightcyan}36 & \cellcolor{highlightcyan}33 \\
 & \textbf{Average} & \textbf{93.00} & \textbf{76.25} & \textbf{49.00} & \textbf{74.50}  & \textbf{34.75}&\textbf{72.75}& \cellcolor{highlightcyan}\textbf{24.25} & \cellcolor{highlightcyan}\textbf{26.00} \\
\bottomrule
  \end{tabular}%
}
\vspace{-3mm}
\end{table*}

\noindent\textbf{Patch objective.} For a fixed distance $D\in\{D_R,D_C\}$, we instantiate the loss introduced in Sec.~\ref{sec:formulation} as
\begin{equation}
\ell_D(p;k,s,\xi,\eta)=-D\bigl(H_p,\operatorname{sg}(H_c)\bigr),
\label{eq:sample_patch_loss}
\end{equation}
where $\operatorname{sg}$ stops gradients through the clean reference. Minimizing this loss maximizes the selected representation discrepancy.

For $z=(k,s,\xi,\eta)$, let $J_z=\partial\mathbf I_p/\partial p$, with observations and patch parameters treated as vectors. The negative gradient of the expected objective is
\begin{equation}
\begin{aligned}
g_D(p)&=-\nabla_p\mathcal L_D(p)\\
&=\mathbb E_z\!\left[J_z^\top
\left(\frac{\partial h_p}{\partial\mathbf I_p}\right)^\top
\nabla_{h_p}D\right].
\end{aligned}
\label{eq:transfer_texture_gradient}
\end{equation}
Here, $D$ is evaluated on the vectorized representations. The compositing Jacobian $J_z^\top$ maps gradients from the two camera views back to the shared patch parameters.

\noindent\textbf{Optimization procedure.} At each update, we sample one state and its associated transformation and action noise for every optimization task. With $B=|\mathcal K_{\mathrm{opt}}|$, the batch objective is
\begin{equation}
\widehat{\mathcal L}_{D,B}(p)=\frac1B\sum_{b=1}^{B}
\ell_D(p;k_b,s_b,\xi_b,\eta_b).
\label{eq:batch_patch_loss}
\end{equation}
We minimize the batch objective using Adam and clip the patch RGB values to $\mathcal P$ after each update. Algorithm~\ref{alg:patch_optimization} summarizes the optimization procedure.

Running Algorithm~\ref{alg:patch_optimization} independently with $D=D_R$ and $D=D_C$ produces the optimized patches $p_D^\star$ for Ours-RMS and Ours-Cosine, respectively.

\begin{table*}[!t]
\centering
\caption{\textbf{Closed-loop evaluation after changing the target object.} Task SR (\%) with success counts, using 60 rollouts per policy and patch. Lower is better. The average gives equal weight to the five policies.}
\vspace{-3mm}
\label{tab:alternate_target}
\small
\setlength{\tabcolsep}{5pt}
\renewcommand{\arraystretch}{1.00}
\resizebox{\textwidth}{!}{%
\begin{tabular}{l*{6}{c}}
\toprule
\rowcolor{gray20}
Policy & UPA-RFAS $\downarrow$ & TAPDreamer $\downarrow$ & DRIFT $\downarrow$ & PASA-KV $\downarrow$ & Ours-RMS $\downarrow$ & Ours-Cosine $\downarrow$ \\
\midrule
FastWAM & 15.00 (9/60) & 5.00 (3/60) & 6.67 (4/60) & 3.33 (2/60) & \cellcolor{highlightcyan}1.67 (1/60) & \cellcolor{highlightcyan}0.00 (0/60) \\
\rowcolor{gray10}
LaWAM & 48.33 (29/60) & 65.00 (39/60) & 20.00 (12/60) & 41.67 (25/60) & \cellcolor{highlightcyan}10.00 (6/60) & \cellcolor{highlightcyan}8.33 (5/60) \\
Cosmos Policy & 38.33 (23/60) & 53.33 (32/60) & 6.67 (4/60) & 55.00 (33/60) & \cellcolor{highlightcyan}1.67 (1/60) & \cellcolor{highlightcyan}1.67 (1/60) \\
\rowcolor{gray10}
OpenVLA-OFT & 25.00 (15/60) & 56.67 (34/60) & 13.33 (8/60) & 58.33 (35/60) & \cellcolor{highlightcyan}3.33 (2/60) & \cellcolor{highlightcyan}0.00 (0/60) \\
$\pi_0$ & 23.33 (14/60) & 56.67 (34/60) & 8.33 (5/60) & 53.33 (32/60) & \cellcolor{highlightcyan}0.00 (0/60) & \cellcolor{highlightcyan}1.67 (1/60) \\
\midrule
\textbf{Average} & \textbf{30.00 (90/300)} & \textbf{47.33 (142/300)} & \textbf{11.00 (33/300)} & \textbf{42.33 (127/300)} & \cellcolor{highlightcyan}\textbf{3.33 (10/300)} & \cellcolor{highlightcyan}\textbf{2.33 (7/300)} \\
\bottomrule
\end{tabular}%
}
\vspace{-4mm}
\end{table*}

\subsection{Native Closed-Loop Transfer}
\label{sec:optimization}

\noindent\textbf{Object-associated deployment.} Given an optimized patch $p_D^\star$, the attacker selects a task object $o\in\mathcal O_k$ and deploys
\begin{equation}
(p_D^\star,T_{k,o}),
\label{eq:deployment_patch}
\end{equation}
where $T_{k,o}$ specifies the task-specific, object-associated placement defined in Sec.~\ref{sec:formulation}. We map the patch onto a non-colliding planar mesh with the same physical size used during optimization. The patch content and world-coordinate pose remain fixed throughout the rollout.

\noindent\textbf{Native scene rendering.} During deployment, native simulator rendering replaces the differentiable compositor $\mathcal C$, and optimization-time EOT is disabled. At policy query $t$, the patched observations are
\begin{equation}
\mathbf I_{p,t}^{\mathrm{native},D}
=\mathcal R_{\mathrm{sim}}\bigl(s_t,p_D^\star,T_{k,o}\bigr),
\label{eq:native_patch_rendering}
\end{equation}
where $s_t$ is the current scene state and $\mathcal R_{\mathrm{sim}}$ renders the patch-containing scene from the policy's cameras. Perspective and occlusion are determined by the current scene geometry and camera poses, while the patch remains stationary in world coordinates.

\noindent\textbf{Closed-loop execution.} The victim policy processes the rendered observations, current proprioception, and original task instruction using its native preprocessing and control procedure. Executed actions update the scene state, and subsequent observations are rendered from the updated scene. No patch updates are performed during execution. For each distance instance, the same optimized patch is reused across tasks, suites, and victim policies; selecting another target object changes its deployment pose $T_{k,o}$, without requiring re-optimization.

 \section{Experiments}
\label{sec:experiments}

\subsection{Experimental Setup}

\noindent\textbf{Dataset.} LIBERO~\cite{liu2023libero} is a simulation benchmark for language-conditioned robotic manipulation with four task families: Spatial, Object, Goal, and Long. We optimize both patch variants using demonstrations from Long and evaluate them on all four families. Spatial, Object, and Goal are held out from patch optimization to assess cross-suite transfer.

\noindent\textbf{Baselines.} We compare against four patch attack methods: UPA-RFAS~\cite{lu2026patch}, TAPDreamer~\cite{lit_2610_06814}, DRIFT~\cite{lit_2608_03207}, and PASA-KV~\cite{lit_2610_39816}, under the same closed-loop evaluation protocol. All baselines are adapted to FastWAM while preserving their core attack formulations and use the same optimization data, patch resolution, physical size, and object-associated placement as OAPA. We also report clean performance and an unoptimized random-patch control on the same scene mesh.

\noindent\textbf{Surrogate and victim policies.} We use FastWAM~\cite{yuan2026fastwam} as the surrogate policy, with white-box access to its frozen parameters for patch optimization. To evaluate transfer across world action models (WAMs), we select LaWAM~\cite{chen2026lawam} and Cosmos Policy~\cite{kim2026cosmos} as victim policies. To further assess transfer across different policy designs, we extend the evaluation to two vision-language-action (VLA) policies, $\pi_0$~\cite{black2024pi0} and OpenVLA-OFT~\cite{kim2025oft}. Each optimized patch is directly deployed to all four victim policies without further optimization. No information about victim policies is used during patch optimization, including weights, architecture details beyond public model names, fine-tuning datasets, recipes, or hyperparameters.

\noindent\textbf{Patch optimization.} We optimize Ours-RMS and Ours-Cosine in separate runs using relative RMS distance and cosine distance, respectively. Each run optimizes a single $50\times50$ RGB patch shared across tasks and views, using 8{,}000 Adam updates at a learning rate of $10^{-3}$. We initialize the patch RGB values uniformly in $[0,1]$. At each update, we sample one demonstration state per optimization task. We use the depth-aware compositing procedure described in Sec.~\ref{sec:physical_rendering}, with the same sampled rotation, brightness, and contrast transformations applied to the patch in both camera views.

\noindent\textbf{Evaluation protocol.} Each suite contains 10 tasks, evaluated from the first 10 official initial states per task, yielding 100 trials per suite and 400 per policy and condition. Patches are rendered on non-colliding square meshes of side length $25\,\mathrm{cm}$ at task-specific, object-associated locations, with poses fixed throughout each trial. Each policy retains its native preprocessing and control protocol. We report LIBERO task success rate (SR), averaged equally across suites; lower SR under attack indicates stronger disruption.

\subsection{Main Results}

\noindent\textbf{White-box attack.} We first evaluate white-box attack performance, with FastWAM serving as both the surrogate and victim policy. Table~\ref{tab:canonical_transfer} shows that Ours-RMS and Ours-Cosine both reduce the average task success rate below all four attack baselines, demonstrating strong white-box attack capability.

\begin{table}[!t]
\centering
\caption{\textbf{First-object completion in failed episodes after target reassignment.} The patch is placed beneath the second target object. Entries report the percentage of failed episodes in which the first target object's task-required manipulation is successfully completed, followed by the corresponding counts in parentheses. Average pools episode counts across the five policies.}
\label{tab:first_object_completion}
\vspace{-1mm}
\small
\setlength{\tabcolsep}{5pt}
\renewcommand{\arraystretch}{1.00}
\begin{tabular}{p{\dimexpr0.32\columnwidth-2\tabcolsep\relax}>{\centering\arraybackslash}p{\dimexpr0.34\columnwidth-2\tabcolsep\relax}>{\centering\arraybackslash}p{\dimexpr0.34\columnwidth-2\tabcolsep\relax}}
\toprule
\rowcolor{gray20}
Policy & Ours-RMS & Ours-Cosine \\
\midrule
FastWAM & 91.53 (54/59) & 95.00 (57/60) \\
\rowcolor{gray10}
LaWAM & 94.44 (51/54) & 100.00 (55/55) \\
Cosmos Policy & 100.00 (59/59) & 98.31 (58/59) \\
\rowcolor{gray10}
\mbox{OpenVLA-OFT} & 87.93 (51/58) & 90.00 (54/60) \\
$\pi_0$ & 63.33 (38/60) & 69.49 (41/59) \\
\midrule
\textbf{Average} & \textbf{87.24 (253/290)} & \textbf{90.44 (265/293)} \\
\bottomrule
\end{tabular}
\vspace{-5mm}
\end{table}

\begin{table}[t]
\centering
\caption{\textbf{Effect of patch placement.} Closed-loop task SR (\%) for Ours-RMS and Ours-Cosine at three object-independent tabletop locations and the default object-associated location (Ours). Locations are defined in the agent-camera view. Each entry aggregates 400 rollouts across four suites. Lower is better.}
\label{tab:placement_ablation}
\small
\setlength{\tabcolsep}{2.5pt}
\renewcommand{\arraystretch}{1.10}
\resizebox{\columnwidth}{!}{%
\begin{tabular}{lcccc}
\toprule
\multicolumn{5}{c}{\textbf{RMS}} \\
\midrule
\rowcolor{gray20}
Policy & Lower-left $\downarrow$ & Center $\downarrow$ & Lower-right $\downarrow$ & \begin{tabular}[c]{@{}c@{}}Object-associated\\(Ours)\end{tabular}\,$\downarrow$ \\
\midrule
FastWAM & 76.00 & 77.75 & 85.25 & \cellcolor{highlightcyan}35.25 \\
\rowcolor{gray10}
LaWAM & 84.25 & 80.25 & 91.25 & \cellcolor{highlightcyan}41.75 \\
Cosmos Policy & 80.25 & 76.75 & 88.00 & \cellcolor{highlightcyan}41.25 \\
\rowcolor{gray10}
OpenVLA-OFT & 86.25 & 75.25 & 85.50 & \cellcolor{highlightcyan}49.25 \\
$\pi_0$ & 72.75 & 50.00 & 77.00 & \cellcolor{highlightcyan}24.25 \\
\midrule
\multicolumn{5}{c}{\textbf{Cosine}} \\
\midrule
\rowcolor{gray20}
Policy & Lower-left $\downarrow$ & Center $\downarrow$ & Lower-right $\downarrow$ & \begin{tabular}[c]{@{}c@{}}Object-associated\\(Ours)\end{tabular}\,$\downarrow$ \\
\midrule
FastWAM & 73.25 & 78.25 & 85.75 & \cellcolor{highlightcyan}32.50 \\
\rowcolor{gray10}
LaWAM & 82.50 & 81.25 & 91.00 & \cellcolor{highlightcyan}36.75 \\
Cosmos Policy & 82.25 & 74.75 & 90.75 & \cellcolor{highlightcyan}38.75 \\
\rowcolor{gray10}
OpenVLA-OFT & 87.00 & 78.00 & 85.50 & \cellcolor{highlightcyan}50.00 \\
$\pi_0$ & 77.00 & 50.50 & 82.50 & \cellcolor{highlightcyan}26.00 \\
\bottomrule
\end{tabular}%
}
\vspace{-4mm}
\end{table}

\noindent\textbf{Black-box transfer.} We transfer each optimized patch to four victim policies without further optimization. Table~\ref{tab:canonical_transfer} shows that both Ours-RMS and Ours-Cosine achieve lower average task success rates than all four attack baselines on every victim policy. For transfer from FastWAM to LaWAM, the clean policy achieves an average SR of 99.00\%, whereas Ours-RMS and Ours-Cosine reduce it to 41.75\% and 36.75\%, respectively. In comparison, the attack baselines retain average SRs between 56.00\% and 88.50\%. Both instances also outperform all baselines in average SR on Cosmos Policy, demonstrating transfer between WAM policies.

The attack also extends beyond Long, the suite used for patch optimization. On the held-out Object suite, Cosmos Policy achieves 100\% clean success, while Ours-RMS and Ours-Cosine reduce SR to 8\% and 7\%, respectively, compared with 21\% for the strongest baseline, DRIFT. These results demonstrate that the optimized patches remain effective under changes in both policy and task suite.

Beyond transfer between WAMs, we further evaluate the patches on OpenVLA-OFT and $\pi_0$. Both instances outperform all baselines in average SR on these VLA policies. On $\pi_0$, Ours-RMS and Ours-Cosine reduce average SR to 24.25\% and 26.00\%, respectively, compared with 34.75\% for the strongest baseline, DRIFT. Together, these results demonstrate the strong cross-suite attack capability and cross-policy black-box transferability of our patches.

\raggedbottom
\subsection{Changing the Target Object During Evaluation}
\label{sec:alternate_target}

\noindent\textbf{Protocol.} We screen all four LIBERO suites for tasks containing two movable task-specified target objects and identify six eligible tasks, all from Long. We relocate each optimized patch from beneath the first target to beneath the second without changing its content or re-optimizing it. Ours-RMS, Ours-Cosine, and all four baselines are evaluated on five policies with 10 episodes per task, totaling 60 episodes per policy and method.

\noindent\textbf{Results.} After relocation beneath the second target object without changing their content, our patches retain strong attack effectiveness. Both Ours-RMS and Ours-Cosine achieve lower task success rates than all four baselines on every policy (Table~\ref{tab:alternate_target}). Across 300 episodes per method, Ours-RMS leaves only 10 successful episodes and Ours-Cosine only seven, corresponding to success rates of 3.33\% and 2.33\%, respectively. These results demonstrate that the same optimized patch remains effective after the associated task object is changed, without object-specific re-optimization.

\noindent\textbf{Object-level failure analysis.} Whole-task failure alone does not establish which object's manipulation is disrupted. We therefore examine whether the first target object's task-required manipulation is completed in failed episodes. Under Ours-RMS, this occurs in 253 of 290 failed episodes (87.24\%); under Ours-Cosine, it occurs in 265 of 293 failed episodes (90.44\%) (Table~\ref{tab:first_object_completion}). Inspection of the rollout videos further shows that failures predominantly occur during manipulation of the second target object, beneath which the patch is placed. Together, these findings support placement-based object-level targeting: the robot usually completes manipulation of the first object, while disruption concentrates on the selected second object.

\subsection{Effect of Patch Placement}
\label{sec:placement_ablation}

\noindent\textbf{Protocol.} We compare object-associated placement with three object-independent tabletop locations: lower-left, center, and lower-right in the reference agent-camera view. For each patch instance, content, physical size, and rendering procedure remain unchanged, with no further optimization. Each placement is evaluated on all five policies across four suites, using 10 episodes per task and 400 episodes per policy and patch.

\noindent\textbf{Results.} Object-associated placement yields lower task success rates than all three object-independent alternatives for both patch instances on every policy (Table~\ref{tab:placement_ablation}). Relative to the most effective object-independent location for each policy, Ours-RMS further reduces SR by 25.75 -- 40.75\% , while Ours-Cosine achieves reductions of 24.50 -- 44.50\%. With patch content and physical size held constant, these results demonstrate that object-associated placement substantially strengthens attack effectiveness across the evaluated policies.

\subsection{Real-World Experiments}
\label{sec:real_world_experiments}

To examine physical applicability beyond simulation, we print the optimized Ours-RMS and Ours-Cosine patches and evaluate them in real-world robot experiments. Both printed patches disrupt the robot's execution of manipulation tasks, demonstrating that their adversarial effects extend beyond simulated observations. The experimental setup, evaluation protocol, and detailed results are provided in the supplementary material.

\FloatBarrier

\section{Conclusion}
\label{sec:conclusion}

We present OAPA, a universal and transferable physical patch framework that selects the task object to disrupt through object-associated placement. OAPA combines physically consistent compositing with final action-expert representation divergence. Evaluations across five policies and four LIBERO suites demonstrate strong attack effectiveness and black-box transfer. Relocating the same patch redirects disruption toward another selected object without re-optimization, revealing a spatially controllable adversarial threat to robotic manipulation.

{\small
\bibliographystyle{ieeenat_fullname}
\bibliography{main,method_references}
}

\end{document}